\documentclass[]{interact}

\usepackage{epstopdf}
\usepackage[caption=false]{subfig}
\usepackage[ruled]{algorithm2e}
\usepackage{multirow}
\usepackage{xurl}
\usepackage[scr]{rsfso}
\usepackage{xcolor}
\usepackage{soul}

\usepackage[natbibapa,nodoi]{apacite}
\theoremstyle{plain}

\theoremstyle{definition}

\theoremstyle{remark}

\begin{document}


\title{Predictive and Prescriptive Analytics for Multi-Site Modelling of Frail and Elderly Patient Services}
\author{
\name{Elizabeth Williams\thanks{CONTACT: E. Williams Email: Williamsem20@cardiff.ac.uk}, Daniel Gartner and Paul Harper}
\affil{School of Mathematics, Cardiff University}
}

\maketitle

\begin{abstract}
Many economies are challenged by the effects of an ageing population, particularly in sectors where resource capacity planning is critical, such as healthcare. This research addresses the operational challenges of bed and staffing capacity planning in hospital wards by using predictive and prescriptive analytical methods, both individually and in tandem. We applied these methodologies to a study of 165,000 patients across a network of 11 hospitals in the UK. Predictive modelling, specifically Classification and Regression Trees, forecasts patient length of stay based on clinical and demographic data. On the prescriptive side, deterministic and two-stage stochastic optimisation models determine optimal bed and staff planning strategies to minimise costs. Linking the predictive models with the prescriptive optimisation models, generates demand forecasts that inform the optimisation process, providing accurate and practical solutions. The results demonstrate that this integrated approach captures real-world variations in patient LOS and offers a 7\% cost saving compared to average-based planning. This approach helps healthcare managers make robust decisions by incorporating patient-specific characteristics, improving capacity allocation, and mitigating risks associated with demand variability. Consequently, this combined methodology can be broadly extended across various sectors facing similar challenges, showcasing the versatility and effectiveness of integrating predictive and prescriptive analytics.
\end{abstract}


\section{Introduction}
Ageing populations put pressure on both the economy and healthcare resources worldwide with medical care expenditures on the rise \citep{Nations2019}. Ageing is one of the most common and well-known risk factors for most chronic diseases \citep{Macnee2014}. According to the \cite{Nations2019}, an elderly person can be defined as 65 years old and over, whilst a frail person is classified as one who is at high risk of falling into dependency as a result of a negative event, such as an injury, fall or disability \citep{Xue2011}. Care for frail and elderly patients has more challenges and barriers in providing care due to a lack of resources or specialised models for care delivery \citep{Heydari2019}. Frail and elderly patients often suffer from multi-morbidity and can take longer to recover in a hospital with more staffing hours and resources required. These patients not only share resources with other patient types but are unique in terms of having their own service specialty within hospitals, known as care of the elderly (COTE). There is often difficulty in grouping these patients together for length of stay (LOS) prediction, as there are many different factors which cause longer lengths of stay. 

To address the challenges to healthcare services caused by ageing populations and the medical complexities of caring for frail and elderly patients, healthcare analytics has proven beneficial. \textcolor{black}{While analytical techniques have been reported to yield up to 25\% reductions in annual healthcare costs \citep{Dash2019}, our study focuses on a specific operational domain-hospital bed and staff capacity planning. Within this context, our integrated predictive and prescriptive models achieve a 7\% cost reduction compared to traditional average-based approaches.}

The recent COVID-19 pandemic has demonstrated the crucial need for analytics to be used and continually developed within the healthcare setting to provide highly beneficial results \citep{Charles2024}. The main challenges faced by healthcare managers are the hundreds of beds and staff to manage, which creates an excessive number of options for decisions \citep{Best2015, Proudlove2007}. Demand and capacity are further complicated because no two patients are exactly alike. Because bed capacity planning typically relies on averages, it fails to take into consideration the stochastic nature of the healthcare industry \citep{ Abdelaziz2012, Harper2002}. This research contributes to the literature by incorporating natural variability through the use of \textcolor{black}{classification and regression trees (CART)}.

\textcolor{black}{This research was motivated by collaboration with a large healthcare network that manages frail and elderly patients across multiple hospitals. Healthcare administrators highlighted significant challenges in bed and staffing capacity planning, where reliance on average-based predictions led to inefficiencies, resource shortages and increased costs. The aim of this study was to develop integrated predictive and prescriptive analytics models that address these operational issues by incorporating patient-specific variability, thereby enhancing planning accuracy and reducing costs.}

To the best of our knowledge, this is the first study to examine the value of linking predictive and prescriptive analytics for the capacity planning of frail and elderly healthcare resources. We approach the problem as follows:
\begin{enumerate}
    \item We begin by utilising CART methodologies with a large dataset consisting of over 165,000 patient records. This analysis focuses on identifying groupings of patients aged 65 and over who share similar clinical and demographical attributes that affect the LOS in hospital. CART was chosen for this analysis because of its ability to handle large datasets effectively and produce easily interpretable visual representations of decision rules. These groupings help us understand the factors that influence LOS and form the basis for our predictive modelling. 
    \item Next, we develop both deterministic and stochastic mixed-integer programming models to plan bed and staffing resources. The deterministic model assumes that future demand is known and fixed, while the stochastic model accounts for uncertainty in patient admissions and LOS by incorporating multiple scenarios into the planning process. These models aim to optimise resource allocation to minimise costs while meeting patient care requirements.
    \item We then integrate the CART-based groupings into our optimisation models. By using the demand forecasts generated from the predictive models, we inform the deterministic and two-stage stochastic optimisation models. This step ensures that the variations and uncertainties captured by the predictive analysis are directly utilised in the resource planning process, leading to more accurate and practical solutions.
    \item Finally, we analyse the value of the stochastic solution (VSS), to assess the benefits of implementing the stochastic model over the traditional deterministic model. VSS measures the improvement in decision-making that results from considering uncertainty in the optimisation process. By comparing the outcomes of both models, we demonstrate the advantages of incorporating stochastic elements into capacity planning, such as better handling of demand variability and improved resource allocation.
\end{enumerate}

\textcolor{black}{The goal of this research is to develop and evaluate decision models that enhance capacity planning for frail and elderly patients in hospital settings. The CART models aim to forecast patient LOS based on clinical and demographic data, facilitating more accurate demand prediction and grouping patients by shared characteristics that influence LOS. These predictions serve as inputs to the optimisation models, which determine optimal bed and staffing levels under both deterministic and stochastic conditions. The integration of these predictive and prescriptive models enables healthcare managers to make informed decisions that minimise costs, accommodate demand variability and ensure the delivery of high-quality patient care.} By understanding the value of predictive and prescriptive analytics, healthcare managers can avoid traps by planning resources based on averages. This, in turn, can lead to better health outcomes and more efficient use of resources.

The remainder of the paper is structured as follows. Section~\ref{relatedwork} provides an overview of current literature in terms of CART analysis with a particular focus on frail and elderly patients. Moreover, deterministic and stochastic optimisation models for hospital bed planning are compared and contrasted with our work. In Section~\ref{methods}, we first introduce CART, followed by the introduction of our deterministic and two-stage stochastic models. Then, we introduce an illustrative example to demonstrate the models' functionality. In Section~\ref{experimental}, we apply our models to real-world data from a network of hospitals, introducing variation into the model by utilising CART predictions for demands. Then, we provide comparisons between the deterministic and two-stage stochastic models by calculating the VSS. We offer a detailed discussion in Section~\ref{discussion}, providing managerial insights, discussing the generalisability of our results and outlining directions for future research. Our paper concludes with Section~\ref{conclusion}.

\section{Related Work}\label{relatedwork}
This section aims to provide insight into existing operational research (OR) literature specifically focusing on hierarchical CART methodologies along with deterministic and stochastic modelling techniques. The first subsection will focus on CART analysis applied to frail and elderly patients and the second subsection will focus on deterministic and stochastic modelling techniques applied to healthcare settings.

Ensuring healthcare facilities possess adequate resources like staff, equipment, and beds to meet patient demand efficiently is essential for effective healthcare operations management \citep{Ardakani2023, Costa2003, Harper2002a}. Within operational research, capacity planning employs predictive modelling and optimisation techniques to anticipate patient flow, regulate hospital bed occupancy, and allocate resources effectively. Previous literature reviews have delved into these methodologies, focusing on their application in optimising healthcare resource utilisation \citep{Chanchaichujit2019, Humphreys2022}. Various mathematical models, including queuing theory and simulation, are utilised to analyse patient arrival patterns, LOS and resource utilisation, enabling healthcare providers to make informed decisions regarding staffing levels, facility expansions, and resource allocation. Capacity planning plays a crucial role in optimising healthcare delivery, minimising wait times, and enhancing patient satisfaction while ensuring cost-effectiveness and resource utilisation \citep{Bhattacharjee2014, Erhard2018, Liping2018, Nasrabadi2020, Grange2024}.

A wide variety of literature reviews have been published within the healthcare OR field. Typically, literature reviews are more focused on specific methods \citep{Fildes2008}, specific problems \citep{Marynissen2019}, or specific patient types \citep{Williams2021}.  \cite{Fildes2008} explored \textcolor{black}{different} forecasting techniques \textcolor{black}{in the OR domain} whilst, \cite{Marynissen2019} focused on different methods but applied to the context of appointment scheduling within hospitals. \cite{Williams2021} presented a literature review highlighting the use of Operational Research and Management Science (OR/MS) methodologies in addressing the challenges associated with the care of frail and elderly patients. These reviews demonstrate a wide variety of research taking place within the healthcare domain. Our research aims to overlap between these reviews, by applying OR methodologies to elderly and frail patient care.

\subsection{CART Analysis Applied to Frail and Elderly}\label{relatedwork-cart}
CART is a machine learning method for constructing prediction models from data. The algorithm constructs a decision tree which is structured hierarchically. The decision tree asks a series of questions that decide groups into which the data is sorted, in the form of binary recursive partitioning, where each node is split into two groups. CART analysis has often been applied to healthcare for predictive analytics, covering a wide range of medical settings. Within frail and elderly care, the literature can be grouped into two subgroups. The first grouping focuses on illnesses typically suffered by these patients, with the second grouping having a more generic medical setting however using elderly and frail as the patient groups of interest.

\subsubsection{Use of CART within Hospitals}
The works of \cite{Byeon2015}, \cite{Ius2020} and \cite{Watanabe2018} demonstrated the use of CART models within hospitals, \textcolor{black}{focusing on age as a predictive factor or limiting the analysis to older populations}. \cite{Byeon2015} developed a CART\textcolor{black}{-based tool to predict} endocrine disorders in the elderly\textcolor{black}{, identifying higher prevalence categories for depression and obesity}. \cite{Ius2020} applied random forests \textcolor{black}{to analyse treatment outcomes, using age as a continuous variable to show it influenced survival, though not as the strongest factor.} \cite{Watanabe2018} applied CART to determine different risk factors of rotator cuff tears between elderly and young patients,\textcolor{black}{find age to be the most significant predictor.} The next category of splitting is then different depending on the age group of the patient. \textcolor{black}{These studies highlight CART’s effectiveness across diverse age-related health conditions.}

\subsubsection{Use of CART across Hospitals and Community Care}
Expanding to hospital and community-wide care, \cite{Kuo2019}, \cite{Lam2019} and \cite{Passmore1993} demonstrate \textcolor{black}{the broader application of CART models.} \cite{Kuo2019} used CART models to develop a system for identifying social frailty in the elderly\textcolor{black}{, incorporating 15 variables (e.g., age, BMI, income)}. Random forests and C5.0 classification models were found to have the highest prediction accuracy of 0.97. \cite{Lam2019} used CART to identify whether frailty would be an indicator of recurrent fallers over the age of 65 in the community. The results showed that certain frailty indexes did have a high predictive ability of recurrent falls, although were only significant in female patients. \cite{Passmore1993} \textcolor{black}{identified that a sickness impact profile (SIP) score was the most influential factor in predicting unplanned hospital admissions among elderly patients. Additionally, the number of medications prescribed was found to influence hospitalisation, suggesting opportunities for preventative measures.}

These six papers demonstrate the success of applying CART to frail and elderly patients when considering both the care setting and the medical condition suffered. 
This research aims to build upon the previous literature in this field by considering how different hospital locations determine a patient's LOS. Additionally, a further variety of data types which may have an impact on patient stays within hospitals, such as radiological data, will be incorporated.

\subsection{Multi-site Deterministic and Stochastic Healthcare Modelling}\label{relatedwork-deterministic}
Deterministic modelling is popular within healthcare settings due to its ease of application, however, traditionally many healthcare services are stochastic in nature \citep{Mandelbaum2018}.

\subsubsection{Deterministic Models}
\cite{Hare2009} and \cite{Segall1992} analysed deterministic modelling within healthcare settings. \cite{Hare2009} developed a deterministic multi-state Markov model to plan for services within home and community care, \textcolor{black}{incorporating five age groups, three of which represent the elderly. }The authors incorporated the changing age and health demographics of the patient groups to determine how service use will change over time. \cite{Segall1992} extended a disaggregated resource allocation model to include a demand constraint, to determine spatial allocation of resources. Applying to 16 acute care hospitals and then to the entire state of Massachusetts, the author was able to determine occupancy rates and demands for hospitals, and specialties and predict the average LOS.

\subsubsection{Stochastic Models}
The works of \cite{Abdelaziz2012}, \cite{Guo2019}, \cite{Levis2004}, \cite{Shehadeh2022} and \cite{Thompson2009} applied stochastic modelling techniques to healthcare settings. \cite{Abdelaziz2012} \textcolor{black}{used a multi-objective stochastic program to determine optimal bed and staffing allocations in 157 public hospitals. The authors considered random demand in order to minimise costs across specialty levels. } \cite{Guo2019} applied logic-based benders decomposition and binary decision diagram-based approaches to optimise surgery scheduling\textcolor{black}{, incorporating uncertainties such as cancellations and procedure durations. These methods successfully generated more robust schedules, reducing cancellations and improving operating room utilisation.} \cite{Levis2004} applied stochastic capacity planning to the pharmaceutical industry\textcolor{black}{, addressing multi-site, multi-period problems with uncertain clinical trial outcomes and customer demand}. The authors provided five examples of planning problems \textcolor{black}{which} are solved with uncertain clinical trial outcomes and customer demand. \cite{Shehadeh2022} used stochastic optimisation for elective surgery scheduling\textcolor{black}{, emphasising that ignoring uncertainties in surgery durations and postoperative recover times results in suboptimal schedules and higher costs.} The paper discusses the complexity of modelling these uncertainties due to limited data, and the conflicting objectives of various stakeholders. \cite{Thompson2009} used Markov decision processes to plan short-term allocation of patients during demand surges. The authors aimed to determine the best patient assignments and reduce the cost of transferring patients.

\subsubsection{Deterministic and Stochastic Models}
\cite{Mestre2015} used location-allocation models with both deterministic and stochastic approaches for the planning and designing of a network of hospitals. The authors aim to inform decision-makers on how to improve access to different healthcare services and specialties whilst minimising costs. \textcolor{black}{The authors} were able to successfully conclude that by including both location and allocation within the first stage, the model was more flexible in terms of hospital network planning, allowing the second stage to incorporate unsatisfied demand and extra capacities. \cite{Restrepo2020} presents a two-stage stochastic model for integrated staffing and scheduling in home healthcare\textcolor{black}{, demonstrating that it reduces under-covering and over-covering costs compared to a deterministic model using average demand.} The study demonstrates that accounting for demand variability through stochastic modelling leads to more robust and cost-effective staffing and scheduling decisions in home healthcare services. \cite{Dehghani2021} proposed both deterministic and stochastic programming models for proactive transshipment in blood supply chains to balance the wastage and shortage of blood units. The deterministic model provided a baseline by assuming known and fixed demand, while the stochastic model accounted for demand uncertainty across a network of hospitals. \textcolor{black}{Using a two-stage stochastic approach with Quasi-Monte Carlo sampling, the study optimised blood orders and hospital transfers, demonstrating significant cost reductions, improved inventory management, and potential savings compared to current and no-transshipment policies.} \cite{Galli2021} used prescriptive analytics for \textcolor{black}{healthcare} inventory management, proposing a novel method that combines machine learning with stochastic optimisation to optimise drug replenishment.  The study demonstrates the effectiveness of this technique in providing robust and cost-efficient inventory solutions tailored to the dynamic and uncertain nature of healthcare environments.

These papers display the variety of both deterministic and stochastic modelling applications within healthcare. These models will be built upon to determine where specific specialties should be located based on demand locations when focusing on a subgroup of patients. The benefits will be determined by using either deterministic or stochastic modelling.

\subsection{Literature Summary}\label{relatedwork-summary}
In conclusion, our search of the literature has revealed gaps in linking predictive and prescriptive analytics to optimise resource allocation under uncertainty in healthcare. While existing studies often use deterministic models, our approach integrates data-driven predictive modelling with stochastic optimisation. We employ CART to forecast demand in a two-stage stochastic program, offering a robust optimisation model. Unlike previous studies focused on single-site models, our work emphasises an integrated network of hospitals, specifically in elderly care. We uniquely optimise staffing and bed capacity jointly and explicitly account for the specialty and acuity mix of frail elderly patients. These aspects, including multi-site planning, predictive forecasting, stochastic optimisation, and a focus on elderly care, constitute our key contributions to the healthcare operations literature.

In terms of the deterministic and stochastic models, our approach builds upon the work of \cite{Maggioni2013}. Our work extends this two-fold: first, by extending the formulation which provides a healthcare example and second, by incorporating CART models into the demand function to create real-world variation in the formulation. We extend this work by providing a healthcare example and incorporating CART models into the modelling formulation.

Second, we research an area of healthcare which has been under-researched, despite the increased demand and medical needs of these patients \citep{Williams2021}.

Finally, with respect to methodology, we provide a different perspective on how this could be modelled. Previous bed and staffing capacity models have focused on neural networks \citep{Kutafina2019}, queuing theoretical approaches \citep{Ghayoomia2022, Vericourt2011, Yankovic2011}, simulation \citep{Amelia2021, Lu2021}. Our methodology builds on traditionally used deterministic methods in healthcare by demonstrating the benefits of using stochastic models by calculating the VSS.

\section{Methods}\label{methods}
This section will discuss the approaches used within this paper, providing the notation used to allow the reader to apply the theories to their own work. There is a large amount of variation in terms of hospital LOS. We have carefully chosen these methods to allow easy incorporation of this variation. CART was chosen over other predictive methodologies because it offers a visual representation. This makes it possible for healthcare practitioners to comprehend and trust the model's output, with the potential for collaboration to create clinically and statistically meaningful groupings. Due to the complexity of our problem, a mixed integer programming approach is used, allowing the LOS variation to be integrated into the bed and staffing planning simultaneously.

\subsection{Classification and Regression Trees}\label{methods-cart}
CART is a non-parametric, supervised machine learning technique that can be used for both classification and regression tasks. It works by recursively partitioning the data into smaller subsets based on predictor variable values. The partitioning process creates a tree-like model of decisions and possible consequences, which are the terminal nodes or leaf nodes of the tree.

One common criterion utilised in the regression side of CART is the mean squared error (MSE) to determine the optimal split at each node \citep{Zanakis2005}, shown in Equation \eqref{Eq:MSE1}.

\begin{equation}\label{Eq:MSE1}
    \text{MSE} = \frac{1}{n}\sum^{n}_{i=1}(Y_{i} - \hat{Y}_{i})^2
\end{equation}
where $n$ is the number of observations, $Y_{i}$ is the actual value and $\hat{Y}_{i}$ is the predicted value.

Algorithm \ref{Alg:Regression} has been formulated to demonstrate the process of regression tree building within the sklearn package within Python.
\begin{algorithm}
\caption{Regression Tree}\label{Alg:Regression}
Determine stopping criteria:\\ $max\_depth, min\_samples\_split, min\_samples\_leaf,$\\$ min\_weight\_fraction\_leaf, max\_leaf\_nodes, $\\$min\_impurity\_decrease$\\
Start with a single node $n$ containing all points. \\
Calculate MSE$^{n}$\\
\While{MSE$^{n} > 0$ \textbf{ or} stopping criterion not met}{
    $k = \text{number of binary splits}$\\
    \For{$a = 1$ to $k$}{
        Calculate MSE$^{n}_{a}$ \\
        $x_{a} = \text{MSE}^{n} - \text{MSE}^{n}_{a}$\\
    }
    Set MSE$^{n} = \max (x_{a})$\\
    \textit{Create two new nodes, n' and n'', and calculate new MSE$^{n}$ for each.}
}
\end{algorithm}

The Gini index is a common criterion used within the classification side of CART and is used to determine the optimal split at each node. The formulation is given in Equation \eqref{eq:Gini}.

\begin{equation}\label{eq:Gini}
    \text{Gini Index} = 1 - \sum_{i=1}^{n}p_{i}^{2}
\end{equation}
where $i$ is the number of classes and $p_{i}$ is the probability of an object that is being classified to a particular class.

The respective algorithm for building a classification tree using the Gini index as the splitting criterion is given in Algorithm \ref{Alg:Classification}.

\begin{algorithm}
\caption{Classification Tree}\label{Alg:Classification}
Determine stopping criteria:\\ $max\_depth, min\_samples\_split, min\_samples\_leaf, $\\$ min\_weight\_fraction\_leaf, max\_leaf\_nodes, $\\$min\_impurity\_decrease$\\
Start with a single node $n$ containing all points. \\
Calculate Gini-Index$^{n}$\\
\While{Gini-Index$^{n} > 0$ \textbf{or} stopping criterion not met}{
    $k = \text{number of binary splits}$\\
    \For{$a = 1$ to $k$}{
        Calculate Gini-Index$^{n}_{a}$ \\
        $x_{a} = \text{Gini-Index}^{n} - \text{Gini-Index}^{n}_{a}$\\
    }
    Set Gini-Index$^{n} = \max (x_{a})$\\
    \textit{Create two new nodes, n' and n'', and calculate new Gini-Index for each.}
}
\end{algorithm}

CART offers several advantages that make it an attractive choice for both classification and regression problems. One of the primary benefits is its interpretability. CART models produce intuitive decision trees that are easy to understand and explain, even for non-experts. The hierarchical structure of the trees provides a clear visualisation of the decision-making process, allowing for straightforward interpretation of the relationships between predictors and the target variable. Healthcare professionals, often not trained extensively in mathematics or statistics, find CART particularly appealing due to its user-friendly interpretability \citep{Bertimas2021}. The simplicity of the decision trees generated by CART allows these professionals to grasp the underlying logic without needing advanced mathematical expertise. This accessibility empowers healthcare practitioners to not only utilise CART effectively but also to confidently communicate its findings and implications to patients and colleagues.

Another significant advantage of CART is its ability to handle non-linear relationships and complex interactions between variables. Unlike linear models, CART can effectively capture intricate patterns and decision boundaries, making it suitable for a wide range of problems involving non-linear relationships. Additionally, CART algorithms perform automatic feature selection during the tree-building process, identifying the most important predictors and reducing the need for manual feature engineering.

CART models are also robust to missing data, as they can handle incomplete datasets by incorporating surrogate splits or imputation techniques during the tree-building process. Furthermore, CART does not require any assumptions about the underlying data distribution, making it a non-parametric and flexible method applicable to a variety of data types and distributions.

\subsection{Deterministic and Two-Stage Stochastic Programming}\label{methods-deterministic}
Within this section the deterministic and stochastic mathematical programs will be discussed, providing the notation used to allow the reader to apply the theories to their own work. The aim of the model is to determine the number of beds and nursing staff required for each specialty within each hospital. The motivation for choosing these resources stems from their pivotal roles in shaping healthcare service delivery. Beds directly influence patient capacity, while nursing staff are instrumental in ensuring high-quality patient care and operational efficiency. Importantly, the allocation of these resources is inherently intertwined, as the number of beds required directly affects the staffing needs, and vice versa. This research extends on the framework used by \cite{Mestre2015} and \cite{Maggioni2010}. 

\subsubsection{General Formulation}
Let us define the two-stage stochastic problem, where a decision-maker takes the decision $x$ of solution space X to minimise expected costs:
\begin{equation}\label{eq:firststage}
    \min _{x \in X} E_{\boldsymbol{\xi}} z(x, \boldsymbol{\xi})=\min _{x \in X}\left\{f_1(x)+E_{\boldsymbol{\xi}}\left[h_2(x, \boldsymbol{\xi})\right]\right\}
\end{equation}
where $x$ are the first-stage decision variables restricted to the set $X \subset \mathbb{R}^{n}$ and $f_{1}(x)$ is the value of the first stage problem. $E_{\boldsymbol{\xi}}$ indicates the expectation with respect to a random vector denoted $\boldsymbol{\xi}$ defined on the probability space ($\Omega$, $\mathscr{A}$, p), with $\Omega \in \mathbb{R}^{n}$ and probability distribution p on the $\sigma$-algebra $\mathscr{A}$.

The function $h_2(x, \xi)$ is the value function of the second stage of the stochastic problem, defined as follows:
\begin{equation}\label{eq:secondstage}
    h_{2}(x,\xi) = \min_{y\in Y (x,\xi)} f_{2}(y:x,\xi)
\end{equation}
where $y$ is the second-stage solution which is restricted to the set $Y \in \mathbb{R}^{n}$. 

Equation \eqref{eq:secondstage} reflects the costs associated with the information being revealed through the realisation of $\xi$ from the random vector $\boldsymbol{\xi}$. The term $\left[h_2(x, \boldsymbol{\xi})\right]$, is known as the recourse function. 

The solution obtained is defined as the `here and now solution' (RP) and is the optimal value of the objective function:
\begin{equation}
    RP = E_{\boldsymbol{\xi}} z(x^{*},\boldsymbol{\xi})
\end{equation}

Equation \eqref{eq:firststage} can be considered where the decision-maker replaces the random variables with their expected values and in turn, solves a deterministic model. This is also known as the expected value \textcolor{black}{(EV)} problem.
\begin{equation}\label{eq:deterministic}
    EV = \min_{x\in X} z(x, \bar\xi)
\end{equation}
where $\bar \xi = E(\boldsymbol{\xi})$, which is the expected value of the random vector $\xi$ and $z$ is the objective value.

\subsubsection{Sets}
The sets used within the deterministic and two-stage stochastic models are pivotal for defining the parameters and variables within the model. Each set serves a distinct purpose. Firstly, $\mathcal{B}$ represents the set of nursing bands, encompassing different skill levels and experiences of nurses. Secondly, $\mathcal{S}$ denotes the set of specialties, where each specialty must be assigned to at least one hospital ($\mathcal{S} \subseteq \mathcal{H}$). Thirdly, $\mathcal{H}$ signifies the set of hospitals, with each hospital being associated with at least one region ($\mathcal{H} \subseteq \mathcal{R}$). Moreover, $\mathcal{R}$ stands for the set of regions, ensuring geographical coverage across the healthcare system. Regions, in this context, could be defined as counties or districts, each comprising of multiple hospitals operating within the same healthcare system organisation. Therefore $|\mathcal{H}| \leq |\mathcal{R}|$. Lastly, $\mathcal{K}$ indicates the set of scenarios, capturing various potential circumstances or conditions. (Table \ref{tab:tssms}).

\begin{table}[h!]
    
    \tbl{The sets used within the two-stage stochastic model where (B, S, H, R, K) represent the maximum number of nursing bands, specialties, hospitals, regions and scenarios, respectively.}{
    \begin{tabular}{ccl}\toprule
        \textbf{Set} & \textbf{Range} &\textbf{Definition} \\\midrule
       $\mathcal{B}$ & b = 1,..., B & \text{Set of nursing bands}\\     
       $\mathcal{H}$ & h = 1,..., H & \text{Set of hospitals} \\
    $\mathcal{K}$ & k = 1,..., K & \text{Set of scenarios}\\
    $\mathcal{R}$ & r = 1,..., R & \text{Set of regions}\\
        $\mathcal{S}$ & s = 1,..., S & \text{Set of specialties} \\\bottomrule
    \end{tabular}}
    \label{tab:tssms}
\end{table}

\subsubsection{Parameters}
Table \ref{tab:tssmp} displays the parameters used within the deterministic and two-stage stochastic models.

\begin{sidewaystable}[htbp]
    \tbl{The parameters used within the two-stage stochastic model where (\(b, s, h, r, k\)) represent the indices of nursing bands, specialties, hospitals, regions and scenarios, respectively.}
   {\begin{tabular}{ll}\toprule
        \textbf{Parameter} & \textbf{Definition}  \\\midrule
            c$^\textnormal{bed, 1st}_{s,h}$& \text{Cost of the first stage beds for specialty $s\in\mathcal{S}$, in hospital $h\in\mathcal{H}$}\\ 
    c$^\textnormal{bed, 2nd}_{s,h}$& \text{Cost of the second stage bed per day for specialty $s\in\mathcal{S}$, in hospital $h\in\mathcal{H}$}\\ 
    c$^\textnormal{staff, 1st}_{b}$& \text{Cost of the first stage staff of band $b\in\mathcal{B}$}\\ 
    c$^\textnormal{staff, 2nd}_{b}$& \text{Cost of the second stage staff of band $b\in\mathcal{B}$}\\ 
    D$_{s,r,k}$ & \text{Demand for each specialty $s\in\mathcal{S}$, arriving from region $r\in\mathcal{R}$, for scenario $k\in\mathcal{K}$}\\ 
    $K_{s,h}$ & \text{Number of beds available to open in each specialty $s \in \mathcal{S}$, in hospital $h \in \mathcal{H}$} \\
    p$_{k}$&  \text{Probability of scenario $k \in \mathcal{K}$}\\
    R$_{s,b}$ &\text{Ratio of nursing staff of band $b\in\mathcal{B}$ to patient for each specialty $s\in\mathcal{S}$}\\ 
    UB$^\textnormal{max, bed, 1st}_{h}$ & \text{Upper bound of the number of beds that are able to be deployed in hospital $h\in\mathcal{H}$ in the 1$^{st}$ stage}\\ 
    UB$^\textnormal{max, bed, 2nd}_{h}$ & \text{Upper bound of the number of beds that are able to be deployed in hospital $h\in\mathcal{H}$ in the $2^{nd}$ stage}\\ 
    UB$^\textnormal{max, staff, 1st}_{b}$ & \text{Upper bound of the number of staff of band $b\in\mathcal{B}$ that can be deployed in the $1^{st}$ stage}\\
    UB$^\textnormal{max, staff, 2nd}_{b}$ & \text{Upper bound of the number of staff of band $b\in\mathcal{B}$ that can be deployed in the $2^{nd}$ stage} \\\bottomrule
    \end{tabular}}
    \label{tab:tssmp}
\end{sidewaystable}

\subsubsection{Decision Variables}
The decision variables introduced in Table \ref{tab:tssmdv} determine the number of beds and nursing staff required for each specialty within each hospital.
\begin{sidewaystable}[htbp]
\tbl{The decision variables used within the two-stage stochastic model where (\(b, s, h, r, k\)) represent the indices of nursing bands, specialties, hospitals, regions and scenarios, respectively.}
    {
    \begin{tabular}{ll}\toprule
        \textbf{Decision Variable} & \textbf{Definition} \\\midrule
         $x^\textnormal{bed}_{s,h} \in \mathbb{N}$ & \text{Number of beds planned in the $1^{st}$ stage for specialty $s\in\mathcal{S}$, in hospital $h\in\mathcal{H}$} \\ 
     $x^\textnormal{staff}_{s,b,h} \in \mathbb{N}$ &\text{Number of staff planned in the $1^{st}$ stage for specialty $s\in\mathcal{S}$, of band $b\in\mathcal{B}$, in hospital $h\in\mathcal{H}$}\\
    $u^\textnormal{bed}_{s,r,h,k} \in \mathbb{N}$ &\text{Number of beds needed in the $2^{nd}$ stage for specialty $s\in\mathcal{S}$, for patients arriving from}\\
   & {region $r\in\mathcal{R}$ in hospital $h\in\mathcal{H}$, for scenario $k\in\mathcal{K}$} \\ 
   $u^\textnormal{staff}_{s,b,h,k} \in \mathbb{N}$ & \text{Number of staff needed in the $2^{nd}$ stage for specialty $s\in\mathcal{S}$, of band $b\in\mathcal{B}$, in hospital $h\in\mathcal{H}$,} \\
   &{for scenario $k\in\mathcal{K}$} \\  \bottomrule 
    \end{tabular}}
    \label{tab:tssmdv}
\end{sidewaystable}

\subsubsection{Mathematical Models}
In this section, we present novel deterministic and two-stage stochastic models derived from the introduced sets, parameters, and decision variables. These models serve as the foundation for our analysis and provide a structured framework for addressing the problem at hand. The deterministic model can be defined as follows:

\begin{equation}\label{eq:det_objective}
    \text{min} \sum_{h\in\mathcal{H}}\sum_{s\in\mathcal{S}}\left(c^\textnormal{bed}_{s,h}x^\textnormal{bed}_{s,h} + \sum_{b\in\mathcal{B}}c^\textnormal{staff}_{b}x^\textnormal{staff}_{s,b,h}\right)
\end{equation}
subject to:
\begin{align}
    \sum_{h\in\mathcal{H}} x^\textnormal{bed}_{s,h}&\geq D_{s,r} & & \forall  s \in \mathcal{S}, r \in \mathcal{R} \label{eq:det_con1}\\
    \sum\limits_{b'\in\mathcal{B}: b'\geq b} x^\textnormal{staff}_{s,b',h} &\geq R_{s,b} \cdot x^\textnormal{bed}_{s,h} & & \forall s \in \mathcal{S}, b \in \mathcal{B}, h \in \mathcal{H}\label{eq:det_con2}\\
     {x_{s,h}^\textnormal{bed}} &\leq K_{s,h} & & \forall s \in \mathcal{S}, h \in \mathcal{H}\label{eq:det_con3}\\
   0 \leq \sum_{s\in\mathcal{S}}{x_{s,h}^\textnormal{bed}} &\leq {UB_h^\textnormal{max, bed}} & & \forall h \in \mathcal{H}\label{eq:det_con4}\\
    0 \leq \sum_{s\in\mathcal{S}}\sum_{h\in\mathcal{H}}{x_{s,b,h}^\textnormal{staff}} &\leq UB^\textnormal{max, staff}_{b}& & \forall b \in \mathcal{B} \label{eq:det_con5}
\end{align}
Objective function~\eqref{eq:det_objective} minimises the cost of deploying beds and staff in each specialty and hospital. Constraints~\eqref{eq:det_con1} ensure the number of beds deployed satisfies the demand. Constraints~\eqref{eq:det_con2} make sure the number of staff deployed to each specialty within each hospital meets the minimum requirements. Constraints~\eqref{eq:det_con3} ensure the number of beds deployed cannot exceed the maximum available specialty beds within each hospital. Constraints~\eqref{eq:det_con4}--\eqref{eq:det_con5} denote the decision variables and their domains.

Similarly, the two-stage stochastic model can be defined as follows:

\begin{multline}\label{eq:sto_objective}
    \text{min} \sum_{h\in\mathcal{H}}\sum_{s\in\mathcal{S}}\left(c^\textnormal{bed, 1st}_{s,h}x^\textnormal{bed}_{s,h} + \sum_{b\in\mathcal{B}}c^\textnormal{staff, 1st}_{b}x^\textnormal{staff}_{s,b,h}\right) 
    \\+\sum_{k\in\mathcal{K}}\sum_{h\in\mathcal{H}}\sum_{s\in\mathcal{S}}p_{k}\left(c^\textnormal{bed, 2nd}_{s,h} u^\textnormal{bed}_{s,h,k}+
   \sum_{b\in\mathcal{B}} c^\textnormal{staff, 2nd}_{b}u^\textnormal{staff}_{s,b,k,h}\right)
\end{multline}

subject to:
\begin{align}
    \sum_{h\in\mathcal{H}} ({x_{s,h}^\textnormal{bed}} +  {u_{s,h,k}^\textnormal{bed}}) &\geq D_{s,r,k}  & & \forall s \in \mathcal{S}, r \in \mathcal{R}, k \in \mathcal{K}  \label{eq:sto_con1}\\
    \sum\limits_{b'\in\mathcal{B}: b'\geq b} {x_{s,b',h}^\textnormal{staff}} &\geq R_{s,b}\cdot{x_{s,h}^\textnormal{bed}} & & \forall s \in \mathcal{S}, b \in \mathcal{B}, h \in \mathcal{H} \label{eq:sto_con2a}\\
    \sum\limits_{b'\in\mathcal{B}: b'\geq b} u_{s,b',k,h}^\textnormal{staff} &\geq R_{s,b} \cdot {u_{s,h,k}^\textnormal{bed}} & &\forall s \in \mathcal{S}, b \in \mathcal{B}, h \in \mathcal{H}, k \in \mathcal{K} \label{eq:sto_con2b}\\
    {x_{s,h}^\textnormal{bed}} &\leq K_{s,h} & & \forall s \in \mathcal{S}, h \in \mathcal{H}\label{eq:sto_con3a}\\
        {u_{s,h,k}^\textnormal{bed}} &\leq K_{s,h} & & \forall s \in \mathcal{S}, h \in \mathcal{H}, k \in \mathcal{K}\label{eq:sto_con3b}\\
       0 \leq \sum_{s\in\mathcal{S}}{x_{s,h}^\textnormal{bed}} &\leq {UB_h^\textnormal{max, bed, 1st}} & &\forall h \in \mathcal{H}\label{eq:sto_con4}\\
   0 \leq \sum_{s\in\mathcal{S}}\sum_{h\in\mathcal{H}}{x_{s,b,h}^\textnormal{staff}} &\leq {UB_{b}^\textnormal{max, staff, 1st}}& & \forall b \in \mathcal{B}\label{eq:sto_con5}\\
    0 \leq  \sum_{s\in\mathcal{S}}{u_{s,h,k}^\textnormal{bed}} &\leq {UB_h^\textnormal{max, bed, 2nd}} & & \forall h \in \mathcal{H}, k \in \mathcal{K}  \label{eq:sto_con6}\\
    0 \leq  \sum_{s\in\mathcal{S}}\sum_{h\in\mathcal{H}}{u_{s,b,k,h}^\textnormal{staff}} &\leq {UB_{b}^\textnormal{max, staff, 2nd}} & & \forall b \in \mathcal{B}, k \in \mathcal{K} \label{eq:sto_con7}
\end{align}

The first sum in the objective function \eqref{eq:sto_objective} is the cost of deploying both beds and staff to specialties within each hospital. The second sum represents the additional beds and staff within the same hospital or a different hospital in the same region. The first constraint,~\eqref{eq:sto_con1}, assures the demand for each specialty and region is met by the number of hospital beds deployed. The demand is dependent on the scenario parameter. Constraints~\eqref{eq:sto_con2a} ensures the number of staff deployed meets the minimum requirements for staff on each specialty ward in the first stage, whilst Constraints~\eqref{eq:sto_con2b} ensures this requirement is met in the second stage. Constraints~\eqref{eq:sto_con3a} and \eqref{eq:sto_con3b} ensure the beds deployed do not exceed the maximum number of beds available for each specialty within each hospital. Constraints~\eqref{eq:sto_con4}--\eqref{eq:sto_con7} denote the decision variables and their domains.

\subsubsection{Evaluation Measures}
Within prescriptive analytics, it is widely recognised that the EV solution can behave poorly in the stochastic domain. Traditional evaluation tests can be carried out in order to determine how each of the EV, RP and EEV performs and determine their robustness. \textcolor{black}{The EEV can be defined as the expected cost when using the solution $\bar x (\bar \xi)$}. \cite{Maggioni2010} discussed four tests to determine the success of stochastic models. For this research, the first method of determining the VSS will be used.

If we let $\bar x (\bar\xi)$ be the optimal solution to Equation \eqref{eq:deterministic}, we can take values and fix these as the first stage, and then allow the second stage of the stochastic model to be performed.

\begin{equation}\label{eq:EEV}
    EEV = E_{\boldsymbol{\xi}} (z(\bar x (\bar \xi),\boldsymbol{\xi}))
\end{equation}

To determine the VSS, the difference between the EEV and RP can be calculated, measuring the expected increase in value from solving the stochastic solution to the simple deterministic one:
\begin{equation}\label{eq:VSS}
    VSS = EEV - RP
\end{equation}

The VSS measures expected loss when using the deterministic solution. If we have hard constraints, the expected cost of the deterministic solution is often $\infty$. Whereas if we use soft constraints, we can make the expected cost arbitrarily bad by setting penalties high using the deterministic solution. If the VSS is large, this could mean the wrong choice of variables has been chosen or the wrong values have been entered.

\subsubsection{Illustrative Example}
To demonstrate the applicability of our proposed model, a small example with fictional numerical values illustrates the optimisation models and key outputs. A small number of scenarios have been included within the second stage to provide a simple illustration of the stochastic programming approach while retaining computational traceability. A dataset of 15 patients demonstrates how the models perform in the application of elderly and frail patient care (Table \ref{tab:patientmatrix}).

\begin{sidewaystable}[htbp]
\tbl{Illustrative Example Patient Dataset with 15 Patients.}{
    \begin{tabular}{lccccccc}\toprule
       \textbf{Patient Number}  & \textbf{Age} & \textbf{Hospital} & \textbf{LOS} & \textbf{Specialty} &\textbf{Admission Method} & \textbf{Admission Source} & \textbf{Frailty Source} \\\midrule
        Patient 1 & 95 & 1 & 5 & COTE & Emergency & Own Home & 3 \\ 
        Patient 2 & 82 & 1 & 3 & COTE & Emergency & Own Home & 2 \\
        Patient 3 & 89 & 1 & 4& T\&O & Emergency & Own Home & 2 \\
        Patient 4 & 87 & 1 & 4 & T\&O & Elective & Own Home & 2 \\
        Patient 5 & 85 & 2 & 3 & COTE & Elective & Transferred & 1 \\
        Patient 6 & 76 & 2 & 1 & COTE & Elective & Transferred & 1\\
        Patient 7 & 71 & 2 & 1 & T\&O & Emergency & Transferred & 1 \\
        Patient 8 & 96 & 1 & 5 & T\&O & Emergency & Own Home & 3 \\
        Patient 9 & 70 & 2 & 1 & COTE & Emergency & Transferred & 1\\
        Patient 10 & 67 & 2 & 1 & T\&O & Elective & Own Home & 1\\
        Patient 11 & 89 & 1 & 4 & COTE & Elective & Transferred & 3\\
        Patient 12 & 70 & 2 & 1 & COTE & Elective & Own Home & 2\\
        Patient 13 & 75 & 2 & 4 & T\&O & Elective & Transferred & 3\\
        Patient 14 & 72 & 2 & 2 & COTE & Elective & Transferred & 3\\
        Patient 15 & 87 & 1 & 5 & COTE & Emergency & Own Home & 2\\\bottomrule
    \end{tabular}}
    \label{tab:patientmatrix}
\end{sidewaystable}
Within the dataset there are two hospitals within the same region, each serving the same two specialties: care of the elderly (COTE) and trauma and orthopaedics (T\&O). We assume there are two nursing staff band levels required on the wards, with differing staff/bed ratios depending on the specialty. Table \ref{tab:DeterministicWorkedExample} shows the parameters and their corresponding values for the deterministic and stochastic models. 

\begin{table}[htbp]
    \tbl{The parameter values that were used within the deterministic and two-stage stochastic model specifically for the illustrative example.}
    {\begin{tabular}{ll}\toprule
       \textbf{Parameters}  & \textbf{Values} \\\midrule
        1$^{st}$ Stage Bed Costs ($c^\textnormal{bed,1st}_{s,h}$) & $\begin{bmatrix} 20 & 30 \\ 30 & 40 \end{bmatrix}$ \\ [0.5cm]
        2$^{nd}$ Stage Bed Costs ($c^\textnormal{bed,2nd}_{s,h}$) & $\begin{bmatrix} 22 & 33 \\ 33 & 44 \end{bmatrix}$\\ [0.5cm]
        Ratio ($R_{s,b}$) &$\begin{bmatrix}0.29&0.14\\
         0.14&0.29\end{bmatrix}$ \\[0.5cm]
         Maximum Specialty Capacity ($K_{s,h}$) &$\begin{bmatrix}
         20 &25  \\ 20 & 25\\
         \end{bmatrix}$\\[0.5cm]
         1$^{st}$ Stage Staff Costs ($c^\textnormal{staff, 1st}_{b}$) &$\begin{bmatrix}\pounds50, \pounds60\end{bmatrix}$ \\ [0.25cm]
         2$^{nd}$ Stage Staff Costs ($c^\textnormal{staff, 2nd}_{b}$) &$\begin{bmatrix}\pounds55, \pounds66\end{bmatrix}$ \\ [0.25cm]
         Upper 2$^{nd}$ bed limit ($UB^\textnormal{max,bed,2nd}_{h,k}$) &  $\begin{bmatrix} 20 & 20 &20\\ 25 & 25 &25 \end{bmatrix}$\\[0.5cm]
         Upper 1$^{st}$ staff limit ($UB^\textnormal{max,staff,1st}_{b}$)& $\begin{bmatrix} 15,25\end{bmatrix}$  \\ [0.25cm]
         Upper 2$^{nd}$ staff limit ($UB^\textnormal{max,staff,2nd}_{b.k}$)& $\begin{bmatrix} 15 & 15 \\ 25 & 25 \end{bmatrix}$\\ [0.5cm]
         Probability of Scenarios ($p_{k}$) & $\begin{bmatrix} 0.4,0.3,0.3\end{bmatrix}$ \\
         
         \bottomrule
    \end{tabular}}
    \label{tab:DeterministicWorkedExample}
\end{table}

The demand from Table \ref{tab:patientmatrix} can be calculated as follows to determine the average daily bed demand (ADBD):
\begin{equation}{
    \text{ADBD}_{s,h} = \text{ALOS}_{s,h} \times \text{ADNA}_{s,h} } \label{eq:averagebeds1}
\end{equation}
where ALOS is equal to the average LOS and ADNA is equal to the average daily number of admissions.
The demand in terms of scenarios and regions can then be calculated:
\begin{equation}{
   D_{s,r} = \text{ADBD}_{s,r} = \sum_{h \in \mathcal{R}} \text{ADBD}_{s,h} } \label{eq:averagebeds2}
\end{equation}
This produces a value of 16.67 and 19.01 for the D$_{0,0}$ and D$_{1,0}$ parameters, respectively. 
For the two-stage stochastic model, a number of scenarios are required. For this example and to demonstrate the model's functionality, we introduce three scenarios which average to the same deterministic demand: Average demand with a probability of 40\%, demand increasing by 20\% with a probability of 30\%, demand decreasing by 20\% with a probability of 30\%.

Therefore the demand matrix $D_{s,r,k}$ can be represented as: 
$\begin{bmatrix}
    [16.66, 19.99, 13.33] \\
     [19.01, 22.80, 15.20]
 \end{bmatrix}$, where the first index refers to the row and the second index refers to the column. In this instance we only have one region, therefore only one column matrix is shown. The third index refers to the column inside the sub-matrix. Table \ref{tab:Results1} displays the optimal decision variables and objective function values for the illustrative example.

\begin{table}[h!]
\tbl{Deterministic and Two-Stage Stochastic Results for the Illustrative Example - Results are Recorded [(beds), (staff)]}
    {\begin{tabular}{l|cccc|c} \toprule
         & \textbf{s=0, h=0} & \textbf{s=0, h=1}  &\textbf{s=1, h=0}  &\textbf{s=1, h=1}  & \textbf{Objective Function Value ($\pounds$)}\\ \midrule
         EV & [(0), (0, 0)] & [(17), (5, 3)] & [(20), (3, 6)] & [(0), (0, 0)] & 2,050\\
        RP & [(20), (6, 3)] & [(1), (1, 1)] & [(24), (4, 8)] &[(0), (0, 0)] &  2,185\\
        EEV & [(4), (2, 1)] & [(17), (5, 3)] & [(24), (4, 8)] & [(0), (0, 0)] & 2,241 \\ \bottomrule
    \end{tabular}}
    \label{tab:Results1}
\end{table}

The results reveal that the deterministic model deploys fewer beds and nursing staff than the stochastic model, leading to an EV solution that costs approximately two-thirds of the RP. The EEV value is calculated by fixing the first-stage decision variables in the two-stage stochastic model with the optimal values from the deterministic model. This then calculates the values for the second-stage variables and the objective function value. The VSS is then calculated to be the difference between the EEV and RP, resulting in a value of $\pounds$56, a saving of 2.5\%, if the stochastic solution were to be implemented. The results also show that the deterministic model is not robust, as the EEV is greater than the RP. This is due to the deterministic model not being able to account for the uncertainty in the demand.

\section{Case Study of a Network of Hospitals in the U.K.}\label{experimental}
In this section, we apply the predictive and prescriptive analytical methodologies to a case study involving a network of hospitals. We aim to demonstrate the practical application and benefits of our approach in a real-world setting. Three years' worth of data covering the period from April 2017 to March 2020 was provided by a large NHS trust in the U.K. To ensure data integrity and avoid skewing from the COVID-19 pandemic, only data preceding the pandemic was utilised. The dataset comprises 165,188 patient records of individuals aged 65 and over, encompassing 29 different specialties and spanning 11 hospital sites. This comprehensive dataset forms the basis of our analysis and modelling efforts, enabling us to derive insights and optimise resource allocation effectively.

\subsection{Results of CART Analysis}\label{experimental-results-cart}
This section discusses the development and results of the CART models. This section will be split into two subsections, the first for regression trees (Section \ref{experimental-results-cart-regression}) and the second for classification trees (Section \ref{experimental-results-cart-classification}). The results of the CART analysis will be used to inform the deterministic and stochastic models.

\subsubsection{Regression Trees}\label{experimental-results-cart-regression}
Regression trees were developed to predict the LOS of patients. Admission method, admission source, age (as a continuous and grouped measure), day, diagnosis, frailty (as a continuous and grouped), hospital, number of scans and specialty were the nine variables used within the regression trees. 

Age and frailty were included as continuous and grouped variables to determine which would be more appropriate for the regression trees. The models were trained on 80\% of the data with a 20\% test set being used to validate the models. 

The `Scikit-learn' package within Python was used to generate the regression trees utilising the `DecisionTreeRegressor' function. The default parameters were utilised within the model with the exception of the `max\_leaf\_nodes' and the `min\_samples\_leaf' parameters. The default parameters have been successfully employed in other studies \citep{Hancock2022, Heyburn2018, Kilincer2023}. These `max\_leaf\_nodes' and the `min\_samples\_leaf' parameters underwent parameter optimisation through a step wise approach. This was to ensure underfitting and overfitting were avoided \citep{Bramer2007}. Additionally, we wanted to set the upper limit of the `max\_leaf\_nodes' to a manageable amount of groupings. The parameters used have been provided within Table 2 in the Supplementary Material and were selected based on the default values and parameter optimisation to determine the largest $R^{2}$ score.

The largest $R^{2}$ score was achieved by grouping age into five-year intervals and using frailty as a continuous measure. This value was calculated to be 34.2\%. Whilst this is a low $R^{2}$ score, it is expected due to the large variation in LOS within the data (417 days). This result shows the model correctly assigns patients to the correct node 34.2\% of the time.

The achieved $R^{2}$ score of 34.2\% by the regression trees in predicting the LOS of patients may appear low at first glance. However, several factors contribute to this result.

Firstly, the dataset exhibits a wide range of LOS values, with a maximum LOS of 417 days. Such variability in LOS can make accurate prediction challenging, as numerous factors, beyond those included in the model, may influence the length of a patient's stay.

Secondly, despite including relevant variables such as admission method, source, diagnosis, and frailty, the predictive power of the model may be limited by the complexity of the LOS prediction task. Some nuances of patient stays may not be fully captured by the chosen variables, leading to inherent limitations in the model's accuracy.

Moreover, while regression trees offer flexibility and interpretability, they may struggle to capture complex relationships or interactions between variables, especially in datasets with high variability like ours. To mitigate this, we carefully optimised model parameters, including `max\_leaf\_nodes' and `min\_samples\_leaf', to balance model complexity and performance.

Furthermore, it's important to highlight that the utility of decision trees extends beyond predictive accuracy alone. Decision trees offer a transparent and interpretable framework for understanding the factors influencing patient LOS. Healthcare practitioners can readily interpret decision trees, making them valuable tools for clinical decision-making and care management.

{While the achieved $R^{2}$ score may seem modest, the visual representation provided by decision trees empowers healthcare practitioners to identify critical decision points and understand the underlying factors contributing to patient LOS. This interpretability can facilitate targeted interventions and resource allocation, ultimately improving patient outcomes and healthcare delivery efficiency.

Therefore, despite the relatively low $R^{2}$ score, the inherent interpretability and actionable insights offered by decision trees make them a valuable asset for healthcare practitioners seeking to optimise patient care pathways and resource utilisation.

\subsubsection{Classification Trees}\label{experimental-results-cart-classification}
Classification trees were also developed to predict patients who were discharged on the same day or admitted overnight. In addition to the nine variables used within the regression trees, month was also included as a predictor, since it was found to be a significant variable within logistic regression analysis. Similar to the regression trees, the model was performed with both age and frailty as a continuous and grouped measure to determine which would be more appropriate. The models were trained on 80\% of the data with a 20\% test set being used to validate the models. The `DecisionTreeClassifier' function was used to develop the classification trees using the `Scikit-learn' package within Python. The parameters used for the `DecisionTreeClassifier' function are shown in Table 2 in the Supplementary Material. The accuracy score was calculated to be 89.9\% when age was grouped and frailty was used as a continuous measure whilst using one minimum sample per leaf and 30 maximum leaf nodes. The groupings determined by the end nodes of the CART models can be used to determine the demand for each specialty within each hospital.

\subsection{Aggregate Results of the Deterministic vs. Stochastic Multi-site Model}\label{experimental-results-deterministic}
This subsection will utilise the average demand for each specialty and each region from the three years' worth of data, and feed this into the deterministic and two-stage stochastic model. \textcolor{black}{All optimisation models were solved using both Microsoft Excel with OpenSolver and Python with the PuLP optimisation library. Python provided significantly faster computational times, demonstrating the efficiency and practical feasibility of the approach for real-world implementation and rapid decision support. In all cases, models were solved to optimality.} The models have been provided open source via GitHub \citep{Williams2023} 

The 11 healthcare locations can be grouped into six different regions, offering 29 specialties across all hospitals. In practice, there are 90 combinations of where specialties can be located, with not all hospitals offering every specialty. Within the NHS, there are different levels of nursing staff, ranging from band two to band eight. The higher the band, the more senior the staff member. Two levels of nursing bands will be used to develop the models (bands five and six). In order to gather costing figures, open source data from \cite{PHS2021} was used. The costing data and parameter values are shown in Table 1 in the Supplementary Material for reference.

In order to meet the demand and satisfy the constraints, the deterministic model utilised the first-stage variables only. The results yielded a yearly cost of $\pounds$904,281. In total, 1,027 beds across different hospitals were deployed with Figure \ref{fig:detresults1} displaying the precise locations of these beds. In order to satisfy demand a total of 414 NHS nurses across a 24-hour period were required.

\begin{figure}[h!]
    \centering
    \includegraphics[width=0.8\columnwidth]{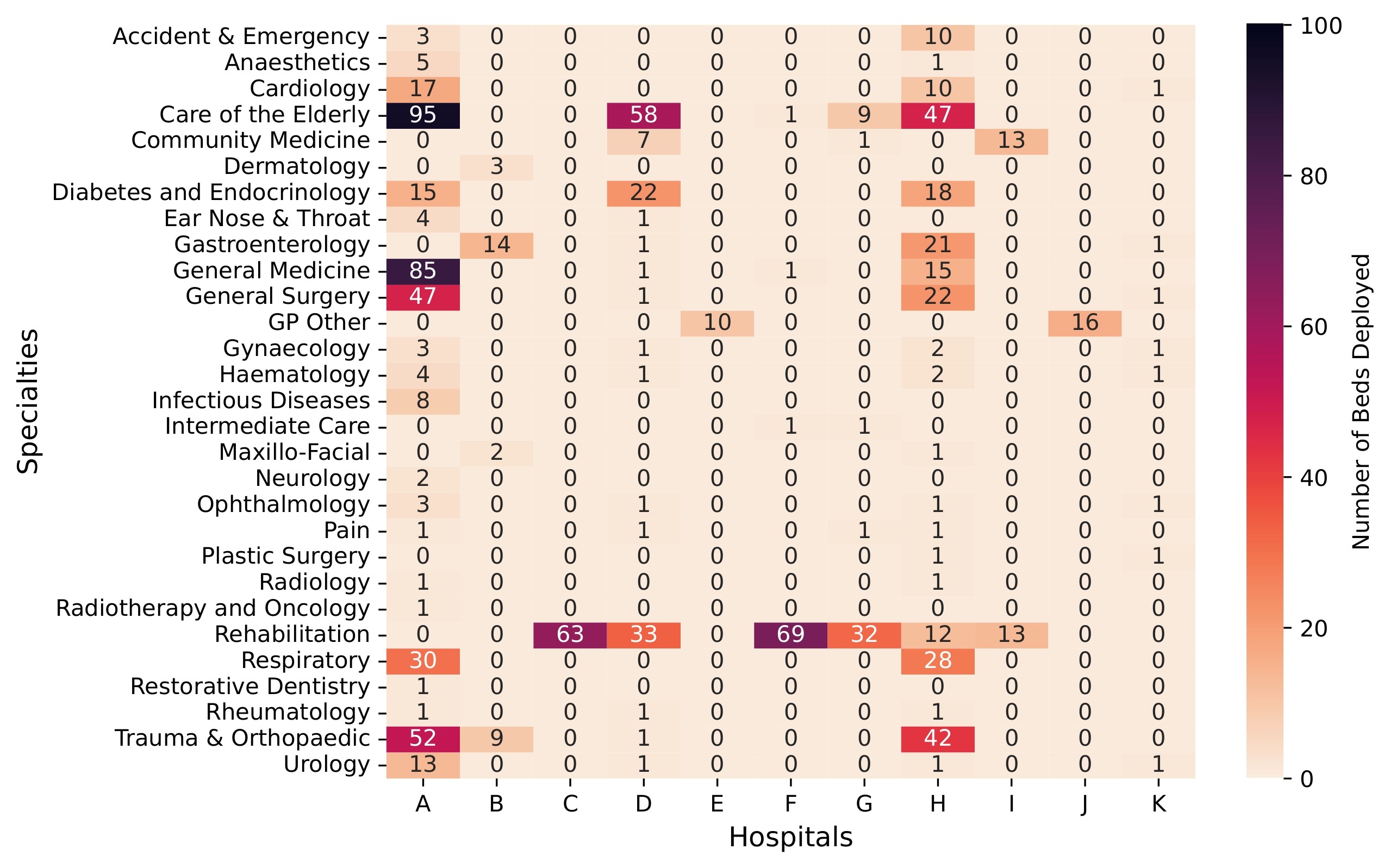}
    \caption{Heatmap of bed locations for each specialty within each hospital for the deterministic model.}
    \label{fig:detresults1}
\end{figure}

The two-stage stochastic model was considered with three different years of data in each scenario, with weightings based on the number of admissions per year. This approach aims to ascertain, drawing insights from past performance, the most effective strategies for future planning and decision-making. Compared to the deterministic model, the stochastic model deployed a higher total number of 1,278 beds across the two stages. The stochastic model also allocated a number of more beds to specialties such as COTE, General Medicine and Rehabilitation. Additionally, the distribution of beds across hospitals differed for certain specialties between the two models, providing more flexibility and adaptability to varying demand patterns in the stochastic approach.
\begin{itemize}
    \item Year one (2017-2018) with a probability of 32.2\%
    \item Year two (2018-2019) with a probability of 34.0\%
    \item Year three (2019-2020) with a probability of 33.8\%
\end{itemize}
The average of all scenarios is equal to the deterministic daily demands. 

The results of the stochastic model yielded a yearly cost of $\pounds$923,828. In total, 941 beds are deployed within the first stage, with a maximum of 337 in the second stage. Similarly, 86 additional nursing staff are deployed across both stages of the model compared to the deterministic counterpart. The location of the 1,278 beds can be seen within Figure \ref{fig:stocHeatmap1}.

\begin{figure}[h!]
    \centering
    \includegraphics[width=0.8\columnwidth]{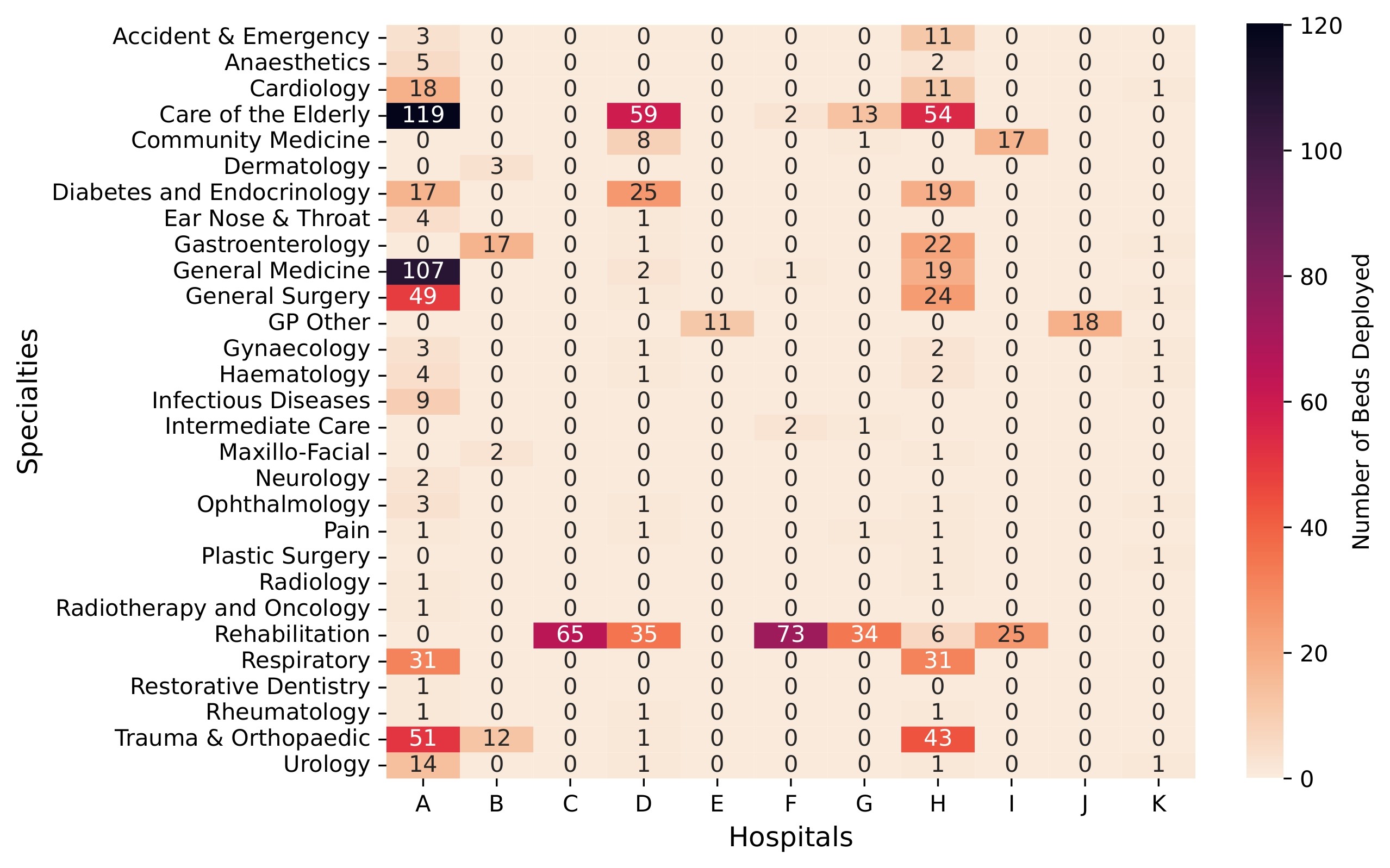}
    \caption{Heatmap of bed locations for each specialty within each hospital for the two-stage stochastic model.}
    \label{fig:stocHeatmap1}
\end{figure}

To provide an understanding of why both models come to different values, we now investigate the robustness of the deterministic model. The VSS was calculated by fixing the results from the deterministic model as the first stage within the two-stage stochastic model. The VSS was calculated to be $\pounds$35,439 over a year period, a saving of $\approx$ 3.8\% (Table \ref{tab:eevdettwostageresults1}). This indicates that the deterministic model is not robust, as the EEV is greater than the RP. The primary reason for this difference is that the deterministic model operates under the assumption of a fixed, known demand, which simplifies the decision making process but fails to incorporate the variability and uncertainty inherent in real-world scenarios.

In contrast, the two-stage stochastic model explicitly considers the uncertainty in demand by allowing for multiple possible future scenarios. This enables more flexible and adaptive decision-making that can better respond to actual demand fluctuations. Consequently, the stochastic model can optimise decisions more effectively, leading to potential cost savings and improved performance.

\begin{table}[h!]
    \tbl{The EV, RP and EEV values for the $x^\textnormal{bed}$, $x^\textnormal{staff}$, $u^\textnormal{bed}$ and $u^\textnormal{staff}$ decision variables and the objective function value.}   
    {\begin{tabular}{cccccc}\toprule 
 & \multicolumn{2}{l}{\textbf{Total Beds}} & \multicolumn{2}{c}{\textbf{Total Staff}} & \multirow{2}{*}{\textbf{Objective Function Value ($\pounds$)}}\\  \cmidrule(lr){2-3} \cmidrule(lr){4-5} 
 & $x^\textnormal{bed}$           & $u^\textnormal{bed}$          & $x^\textnormal{staff}$         & $u^\textnormal{staff}$         \\ \midrule
       EV & 1,026 & - &  414 & - & 904,281 \\ \midrule 
      RP & 941 &  337 & 346 & 154 & 923,828 \\ \midrule
      EEV & 1,026 & 186 & 414&  104 &  959,267 \\\bottomrule
    \end{tabular}}
     \label{tab:eevdettwostageresults1}
\end{table}

\subsection{Results Broken Down by Specialty}\label{experimental-results-specialty}
In this subsection, we break down the results by specialty to understand why the deterministic and stochastic models yield different outcomes. 

The variation in demand, especially for high-impact specialties such as COTE, plays a significant role in these differences. The COTE specialty often has high variability in patient demand and longer LOS, making it a critical factor in resource allocation.

To illustrate, consider the allocation of beds for the COTE specialty across all hospitals. The deterministic model, which does not account for variability in demand, assigns a fixed number of 210 beds for COTE (Figure \ref{fig:detresults1}). This fixed allocation is based on average demand and does not adjust for peak periods, potentially leading to shortages during times of high demand and under-utilisation during low-demand periods. In contrast, the stochastic model incorporates variability and dynamically adjusts resource allocation to better match real-time demand fluctuations. As shown in Figure \ref{fig:stocHeatmap1}, the stochastic model allocates up to 247 beds for COTE when necessary, reflecting its flexibility in responding to peaks in demand. This dynamic adjustment capability allows the stochastic model to efficiently manage resources, reducing the likelihood of shortages and improving overall service delivery.

The stochastic model’s ability to handle variability results in several benefits. Although it requires a higher initial deployment of resources, such as additional beds, this approach ultimately leads to cost savings. By preventing bed shortages and ensuring that patient care needs are met during peak times, the stochastic model minimises the need for expensive emergency measures and improves patient outcomes. This proactive allocation strategy exemplifies the model's superiority in handling high-impact specialties with significant demand variability.

Overall, the comparison between deterministic and stochastic models highlights the importance of incorporating demand variability into resource planning, particularly for specialties with high impact and variability like COTE. By dynamically adjusting resource allocation, the stochastic model demonstrates its effectiveness in optimising resource use and enhancing the healthcare system's responsiveness to patient needs.

\subsection{Results of Linking the Two Paradigms}\label{experimental-results-linking}
To explore the relationship between patient characteristics and LOS within the context of decision trees, we investigated multiple methods that can be employed to derive insights from both classification and regression tree results. These methods offer different perspectives on how patient attributes and healthcare facility factors contribute to LOS prediction.

The first method involved calculating the number of patients of each specialty and determining the overall average LOS for each end node. This approach provides a broad overview of LOS patterns across different patient populations and allows for the identification of specialties with particularly high or low LOS averages.

The second method utilised each end node and considered the specific LOS for each specialty and hospital within the node. By aggregating LOS data at this granular level, we obtained a more detailed understanding of the factors influencing LOS within specific patient groups and healthcare facilities.

Moreover, our analysis demonstrated that employing 30 end node groups, as opposed to traditional methods, offers significant benefits in capturing nuanced relationships between patient characteristics, healthcare settings, and LOS prediction. This approach provides a finer granularity of analysis, enabling more precise identification of factors influencing LOS and facilitating targeted interventions for improved patient outcomes.

Moving forward, we will focus on exploring utilising the classification tree with specific LOS for each node, in more detail to demonstrate its applicability in uncovering nuanced relationships within the British healthcare context. This deeper exploration will shed light on the potential insights that can be gleaned from decision tree analyses in healthcare research and clinical practice.

The process for generating demands based on these specific LOS values is depicted in Equation \eqref{eq:treedemand4}, which accounts for both the number of patients and their LOS within each specialty and hospital.

\begin{equation}\label{eq:treedemand4}
    D_{s,r} = \sum\limits_{h \in r} D_{s,h} = \frac{\text{Number of Patients}_{s,h}\times{\text{Specific LOS}_{s,h}}}{\text{Total Number of Days in Dataset}}
\end{equation}

These refined demand generation processes were used to understand how CART can be linked to both deterministic and two-stage stochastic models, with the results presented in Table \ref{tab:Results9}. Additionally, the EEV was computed to assess the VSS within the context of the models' performance.

\begin{table}[h!]
\tbl{The EV, RP and EEV values for the $x^\textnormal{bed}$, $u^\textnormal{bed}$, $x^\textnormal{staff}$, $u^\textnormal{staff}$ decision variables and objective function using the classification tree and the specific LOS across all three years.}
   {\begin{tabular}{cccccc}\toprule
 & \multicolumn{2}{l}{\textbf{Total Beds}} & \multicolumn{2}{c}{\textbf{Total Staff}} & \multirow{2}{*}{\textbf{Objective Function Value ($\pounds$)}}\\ \cmidrule(lr){2-3} \cmidrule(lr){4-5}
 & $x^\textnormal{bed}$           & $u^\textnormal{bed}$          & $x^\textnormal{staff}$         & $u^\textnormal{staff}$         \\\midrule
    EV      &  1,002 & - &422  & - & 840,245 \\ \midrule
    RP & 911 &  343& 348 & 162 &  862,155 \\ \midrule
    EEV & 1,002& 185 & 422 & 98 & 894,198\\\bottomrule
    \end{tabular}}
    
    \label{tab:Results9}
\end{table}
\textcolor{black}{Calculating the VSS from the results in Table \ref{tab:Results9}, produces a saving of $\pounds$32,0423 per day (3.7\%). Although the calculated VSS is a smaller value compared to the VSS from average based planning, the specific LOS model results in a lower overall EEV, with a 7\% cost difference between the two EEV values.} 
The model suggests deploying fewer beds and nursing staff in the first stage when comparing the RP to the EEV results, however, given the uncertainties of the second stage, it is more cost-efficient to maintain flexibility with fewer resources initially, resulting in long-term savings and improved cost efficiency.

The comparison between the deterministic and CART-linked models reveals significant advantages associated with integrating CART methods into healthcare planning processes. Firstly, the CART approach offers enhanced granularity and insight into LOS prediction compared to traditional deterministic models. By considering specific LOS values for each specialty and hospital within each node, the CART model provides a more nuanced understanding of the factors influencing LOS. This finer level of detail enables healthcare planners to tailor interventions and allocate resources more effectively, addressing the diverse needs of different patient populations and healthcare facilities.

Moreover, the CART-linked models present opportunities for cost savings through more robust decision-making. By leveraging predictive analytics and historical data, the CART model can identify patterns and trends in LOS that may not be apparent in traditional deterministic approaches. This enables healthcare planners to make more informed decisions about resource allocation, optimising bed capacity, staffing levels, and other resources to meet patient demand more efficiently. As a result, healthcare organisations can potentially reduce operational costs while improving patient outcomes and satisfaction.

The VSS savings observed in the CART-linked models further emphasise the potential value of leveraging CART techniques in healthcare resource management. The VSS analysis quantifies the potential cost savings associated with incorporating uncertainty into decision-making processes, highlighting the importance of considering both historical data and predictive analytics in strategic planning. By accounting for variability and uncertainty in patient demand and LOS, the CART-linked models enable healthcare organisations to make more resilient and adaptable decisions, better positioning them to respond to changing healthcare needs and environmental factors.

Overall, the comparison of results between the deterministic and CART-linked models underscores the transformative potential of CART methods in healthcare planning. By providing enhanced granularity, insight, and cost savings opportunities, CART techniques offer a powerful tool for optimising resource allocation, improving operational efficiency, and ultimately enhancing patient care delivery in healthcare settings.

\section{Discussion}\label{discussion}
In this section, we synthesise the key findings of our research and outline their implications for healthcare resource planning. By linking predictive and prescriptive analytics, our research offers a comprehensive approach to optimising hospital bed and staffing allocations, particularly for frail and elderly patients. We will discuss the practical applications of our models, examine their limitations, and suggest directions for future research. Our goal is to provide actionable insights and recommendations that can enhance decision making processes in healthcare management.

The integration of predictive and prescriptive analytics in the context of healthcare resource planning, specifically for frail and elderly patients, presents a robust approach to enhancing decision-making processes. Our research demonstrates the effectiveness of combining CART analysis with deterministic and stochastic programming models. By analysing over 165,000 patient records, CART analysis provided insights into patient groupings based on clinical and demographic attributes affecting the length of stay (LOS). These insights were crucial in developing our predictive models, which in turn informed the resource allocation strategies in our optimisation models.

The deterministic and stochastic models we developed address both the certainty and uncertainty inherent in healthcare demand. The deterministic model, while straightforward, operates under the assumption of fixed future demand, which is rarely the case in real-world settings. In contrast, the stochastic model incorporates multiple scenarios to account for the variability in patient admissions and LOS, offering a more flexible and realistic planning tool. The integration of CART-based predictions into these models ensures that variations and uncertainties are considered, leading to more accurate and practical solutions.

Our findings underscore the value of the stochastic approach. The VSS analysis revealed that incorporating uncertainty into the optimisation process significantly improves resource allocation, reducing costs while maintaining or enhancing patient care quality. This is particularly relevant in the healthcare sector, where demand is highly unpredictable, and the cost of over- or under-estimating resources can be substantial.

\subsection{Bridging the literature gap}
The literature search of Section \ref{relatedwork-summary} has revealed gaps in linking predictive and prescriptive analytics to optimise resource allocation under uncertainty in healthcare. Current literature relies heavily on deterministic optimisation models for healthcare planning, which fail to capture inherent variability. We bridge this gap by integrating data-driven predictive modelling with stochastic optimisation. Specifically, we use CART to generate demand inputs for a two-stage stochastic program, creating a robust optimisation model. This combination of predictive and prescriptive analytics differentiates our work from existing literature. Additionally, our focus on an integrated network of hospitals providing elderly care contrasts with single-site models that dominate previous studies. Furthermore, our joint optimisation of staffing and bed capacity is unique compared to other papers that optimise these decisions separately. Finally, the explicit incorporation of specialty and acuity mix for frail elderly patients differs from previous studies. 

Moreover, our contribution includes the development of a novel model and a comprehensive process/workflow that evaluates the quality of different models. This process ensures rigorous assessment of model performance, including predictive accuracy and optimisation effectiveness, by employing relevant metrics and validation techniques. This systematic approach enhances the reliability and applicability of our findings in real-world healthcare settings, marking a significant advancement in the field.

\subsection{Limitations}
Despite the promising results, our research has several limitations that warrant consideration. Firstly, the CART analysis, while powerful, has inherent limitations. The method can be prone to overfitting, especially when dealing with highly complex datasets. Although we employed techniques to mitigate this risk, such as pruning and cross-validation, the potential for overfitting remains a concern.

Furthermore, the stochastic model's accuracy is heavily dependent on the quality of the input data and the scenarios generated. Inaccurate or biased data can lead to suboptimal resource allocation decisions. While our research utilised robust data and scenario generation techniques, the ever-changing nature of healthcare demand means that continuous updates and validations of the model are necessary.

Lastly, the computational complexity of the stochastic model can be a barrier to its practical implementation. Solving such models requires significant computational resources and time, which may not be feasible for all healthcare institutions, particularly those with limited resources. Future research could focus on developing more efficient algorithms and exploring the use of advanced computing techniques, such as parallel processing, to enhance the model's practicality.

In conclusion, the results from our innovative approach bridging the world of predictive analytics with the prescriptive analytics paradigm provides valuable insights into the application of predictive and prescriptive analytics in healthcare resource planning, these limitations highlight the need for cautious interpretation of the results and suggest directions for future research. Addressing these limitations will be crucial in refining the models and ensuring their broader applicability and effectiveness in diverse healthcare settings.

\subsection{Managerial Insights and Generalisability of the Results}\label{experimental-results-generalizability}
The results discussed can provide guidance and recommendations for enhancing healthcare bed and staffing allocations. One key takeaway from the models is the clear demonstration of the drawbacks of planning solely based on averages. \textcolor{black}{Specifically, the model outputs reveal that using average LOS-based demand predictions yields an EEV that is 7\% higher than when using patient-specific LOS predictions. This cost difference amounts to a potential saving of $\pounds$34,425 per day by accounting for patient-level variability. Additionally, the VSS for the specific LOS model demonstrates that flexibility and scenario-based planning reduce daily operational costs by $\pounds$32,042, further underscoring the financial and resource management benefits of avoiding average-based approaches.} This eye-opening insight has underscored the importance of adopting more sophisticated and dynamic approaches to resource planning, steering the health board away from potential pitfalls in their decision-making process. Perhaps the most impactful aspect of this project lies in its utilisation of predictive modelling, specifically within the healthcare domain. For a healthcare provider who is accustomed to simpler average models, this research has showcased the true potential of mathematical modelling, revealing its power in unravelling complexities, optimising operations, and delivering data-driven insights into healthcare planning. The predictive models developed in this research were trained on data collected under the existing bed and staffing allocations, aiming to capture patterns in patient LOS and demand. While acknowledging the potential for endogeneity in demand generation, particularly in elective specialties, we argue that the observed demand patterns are largely exogenous to the staffing and bedding decisions. This consideration regarding the exogeneity of demand in relation to staffing and bedding decisions should be taken into account when applying the developed predictive models to other healthcare settings or similar problems.

\textcolor{black}{For practitioners to adopt these models effectively, several steps are necessary. First, robust data collection systems must be in place to capture real-time patient information, including clinical and demographic data for predictive modelling. Second, user-friendly decision-support tools need to integrate predictive outputs with optimisation algorithms, allowing healthcare managers to simulate scenarios and adjust resource plans dynamically. Third, training for healthcare decision-makers is essential to interpret model outputs and apply them in policy and operational contexts. Lastly, institutional support for integrating predictive and prescriptive analytics into routine planning processes can facilitate sustained improvements in efficiency and patient care.}

Generalising results is a critical aspect of research that helps to ensure that the findings of a study are relevant and applicable beyond the specific context in which they were obtained. This makes it possible to guarantee that the research will be beneficial and instructive for other academics, professionals, and policymakers who could be working in other locations or with various populations. The deterministic and two-stage stochastic equations are able to be applied to any healthcare scenario. Whilst this research particularly focused on frail and elderly patients, due to the changing population demographics within the area, the equations can be extended to other age groupings. The benefit of using CART models is that researchers and clinicians can apply the theory to their own patient types and identify distinctive homogeneous clusters of patient features. As time passes and the demographic of patients changes, these models can be rerun to determine new patient clusters. The user can choose the number of hospitals in each region and the range of specialties they may provide because of the equations' structure, which allows the models to be adjusted to fit any size region. Whilst these models were run with three levels of nursing bands, these can be increased or decreased to suit the user. Additionally, if decision-makers wanted to determine the needs for other hospital resources such as ventilators, these could be easily added to the model. The models are adaptable and reliable to suit a variety of healthcare situations.

\subsection{Future work}
This work could be extended in several ways based on the required application. One avenue could involve delving into causal models to establish deeper analytical frameworks for understanding the relationships between patient attributes and LOS. Additionally, exploring online algorithms for real-time resource management offers a promising avenue for dynamic allocation strategies, addressing day-to-day or specific time-to-time decisions in healthcare settings, especially considering factors such as hospital utilisation and seasonal demand changes. Implementing reinforcement learning strategies based on Markov decision processes could further enhance resource allocation approaches, particularly in managing dynamic patient demands and optimising resource utilisation. Furthermore, characterising the trade-offs between different methodological approaches, such as the two-stage stochastic program and alternative models like infinite-horizon dynamic programming, will provide deeper insights into methodological choices and their implications for healthcare resource management. Additionally, conducting a comprehensive sensitivity analysis to identify which parameters and modelling assumptions have the biggest potential impact on cost savings and resource optimisation will allow for the prioritisation of refinements aimed at further improving the predictive and prescriptive models.

\section{Conclusion}\label{conclusion}
By linking predictive and prescriptive analytics, decision-makers can obtain a comprehensive view of their data and use it to make better decisions. For example, if predictive analytics indicates there is a high likelihood of a certain event occurring in the future, prescriptive analytics can recommend specific actions that can be taken to mitigate the risk or take advantage of the opportunity. Furthermore, this integration can also allow decision-makers to continuously improve their decision-making processes over time. By tracking the effectiveness of their decisions and making adjustments based on new data and insights, they can optimise their operations and achieve better outcomes.

This research has discussed how predictive and prescriptive analytics could be used in combination for efficiently planning hospital specialty beds and staffing requirements for a network of hospitals in the U.K. By comparing the regression tree and classification results to the averages, it allowed differences to be determined and validation of the linked methods to take place. The results showed regression trees produced closer results to the averages. The validation of these regression trees paves the way for more complex scenario analysis to be able to take place.

\section*{Acknowledgements}
The authors would like to acknowledge the generous support and contributions from NHS staff in the Aneurin Bevan University Health Board as well as the financial contribution from the Knowledge Economy Skills Scholarship (KESS2) and the Leadership Engagement Acceleration \& Partnership (LEAP). KESS2 is a pan-Wales higher level skills initiative led by Bangor University on behalf of the Higher Education sector in Wales. It is part funded by the Welsh Government‘s European Social Fund (ESF) convergence programme for East Wales. LEAP is an EPSRC Digital Health Hub with award reference number EP/X031349/1.

\section*{Compliance with Ethical standards}
This article does not contain any studies with human participants or animals performed by any of the authors.

\section*{Disclosure statement}

The authors have no competing interests to declare that are relevant to the content of this article.


\bibliographystyle{apacite}
\bibliography{references}

\newpage
\appendix
\section{List of Patient Attributes}

\begin{sidewaystable}[htpb]
    \tbl{List of patient attributes within the dataset used for analysis.}    
    {\begin{tabular}{llll}\toprule
        {{\textbf{Attribute}}} & { {\textbf{Data type}}} & { {\textbf{Distinct attribute values or bins}}} & { \textbf{Documentation}} \\         \midrule
        {Admission Date} & {Ordinal} & {1096 (e.g. 01/04/2017)} &Upon admission \\
        {Admission Method} & {Nominal} &  {17 (e.g. Elective waiting list)} &Upon admission \\
        {Admission Source} & {Nominal} & {26 (e.g. Usual place of residence)} &Upon admission \\
        {Admission Time} & {Continuous} & {1440 (\{hh:mm\})} &Upon admission \\
        {Borough} & {Nominal} & {174 (e.g. Newport LHB, Monmouthshire LHB)} &Upon admission \\
        {Date of Birth} & {Ordinal} & {12037 (e.g. 01/01/1940)} &Upon admission \\
        {Diagnosis} & {Nominal} &{2758 (e.g. Fracture of neck of femur, Congestive heart failure)} &Upon admission \\
        {Discharge Date} &Ordinal&1154 (e.g. 01/04/2017)& On discharge \\
        {Discharge Destination} &{Nominal} &{26 (e.g. Death, Own home, Patient transfer within same health board/trust)} & On discharge \\
        {Discharge Time} &Continuous&1306 (\{hh:mm:ss\}) & On discharge \\
        {Hospital} & {Nominal} & {14 (e.g. Chepstow Community Hospital)} &Upon admission \\
        {NHS Number} & {Nominal} & {66251 (e.g. 4900000000)} &Upon admission \\
        {Postcode} & {Nominal} &{13819 (e.g. CF72 8XR)} &Upon admission \\
        {Registered GP} & {Nominal} & {1313 (e.g. G9041668)} &Upon admission \\
        {Registered GP Practice} & {Nominal} & {618 (e.g. W93012)} &Upon admission \\
        {Scan Attendance Date} & {Ordinal}& {1097 (e.g. (01/04/2017)} & While admitted \\
        {Scan Attendance Time} &{Continuous}& {11417 (\{hh:mm:ss\})}&While admitted \\
        {Scan Exam} &{Nominal}& {293 (e.g. CT Neck and thorax)}&While admitted \\
        {Scan Exam Code} &{Nominal}& {295 (e.g. XCHES, XABDO, CSKUH)} &While admitted \\
        {Scan Procedure Code} &{Nominal}& {16 (e.g. R, CT, MR)} &While admitted \\
        {Scan Requested Date} & Ordinal & {1090 (e.g. (01/04/2017)}&While admitted \\
        {Scan Specialty Code} & {Nominal} & {37 (e.g. Gastro, Neuro)} &While admitted \\
        {Specialty} & {Nominal} & {30 (e.g. Care of the Elderly, Neurology)}&Upon admission \\
        \bottomrule

    \end{tabular}}
    \label{tab:Attributes}
    \end{sidewaystable}

\newpage


\end{document}